\documentclass[accepted]{uai2023} 

\usepackage[american]{babel}

\usepackage{natbib} 
\usepackage{mathtools} 
\usepackage{booktabs} 
\usepackage{makecell}
\usepackage{tikz} 

\usepackage{subcaption}
\usepackage{pgfplots}

\usetikzlibrary{arrows.meta}
\usetikzlibrary{backgrounds}
\usepgfplotslibrary{patchplots}
\usepgfplotslibrary{fillbetween}
\pgfplotsset{%
    layers/standard/.define layer set={%
        background,axis background,axis grid,axis ticks,axis lines,axis tick labels,pre main,main,axis descriptions,axis foreground%
    }{
        grid style={/pgfplots/on layer=axis grid},%
        tick style={/pgfplots/on layer=axis ticks},%
        axis line style={/pgfplots/on layer=axis lines},%
        label style={/pgfplots/on layer=axis descriptions},%
        legend style={/pgfplots/on layer=axis descriptions},%
        title style={/pgfplots/on layer=axis descriptions},%
        colorbar style={/pgfplots/on layer=axis descriptions},%
        ticklabel style={/pgfplots/on layer=axis tick labels},%
        axis background@ style={/pgfplots/on layer=axis background},%
        3d box foreground style={/pgfplots/on layer=axis foreground},%
    },
}

\usepackage{siunitx}
\usepackage{amsmath}
\usepackage{amssymb}
\usepackage{bbm}
\usepackage{cleveref}

\usepackage{algorithm}
\usepackage{algorithmicx}
\usepackage[noend]{algpseudocode}

\newsavebox\CBox
\def\mathBF#1{\sbox\CBox{$#1$}\resizebox{\wd\CBox}{\ht\CBox}{$\mathbf{#1}$}}

\title{Conditional Deep Generative Models for Belief State Planning}

\author[1,*]{Anthony Corso}
\author[2,*]{Antoine Bigeard}
\author[1]{Mykel J. Kochenderfer}
\affil[1]{%
    Aeronautics and Astronautics\\
    Stanford University\\
    Stanford, California, USA
}
\affil[2]{%
    Institute for Computational \& Mathematical Engineering (ICME)\\
    Stanford University\\
    Stanford, California, USA
}
\affil[*]{%
Equal contribution
}
    
\begin{document}
\maketitle

\begin{abstract}
  Partially observable Markov decision processes (POMDPs) are used to model a wide range of applications, including robotics, autonomous vehicles, and subsurface problems. However, accurately representing the belief is difficult for POMDPs with high-dimensional states. In this paper, we propose a novel approach that uses conditional deep generative models (cDGMs) to represent the belief. Unlike traditional belief representations, cDGMs are well-suited for high-dimensional states and large numbers of observations, and they can generate an arbitrary number of samples from the posterior belief. We train the cDGMs on data produced by random rollout trajectories and show their effectiveness in solving a mineral exploration POMDP with a large and continuous state space. The cDGMs outperform particle filter baselines in both task-agnostic measures of belief accuracy as well as in planning performance. 
\end{abstract}

\section{Introduction}\label{sec:intro}
Partially observable Markov decision processes (POMDPs) can model a wide range of applications such as robotics~\citep{pineau2007pomdp}, autonomous vehicles~\citep{wray2021pomdps}, and, more recently, subsurface problems such as ground water~\citep{wang2022sequential}, carbon capture and storage~\citep{corso2022pomdp}, and mineral exploration~\citep{mern2023intelligent}. A major challenge for solving POMDPs with high dimensional states is the difficulty of accurately representing the belief, which is the probability distribution over the state of the system given the history of actions and observations.  In this paper, we propose a novel approach to address this challenge using conditional deep generative models (cDGMs) to represent the belief.

\begin{figure}
\begin{subfigure}[b]{\linewidth}
    \centering
    \begin{tikzpicture}

\tikzstyle{simplecircle} = [circle, minimum width=1cm, minimum height=1cm, draw=black]

\node (1) at (0,0) [simplecircle] {};

\filldraw[color=black, fill=black](0.2,0.2) circle (0.08cm);
\filldraw[color=black, fill=black](-0.2,0.2) circle (0.08cm);
\filldraw[color=black, fill=black](0,0) circle (0.08cm);
\filldraw[color=black, fill=black](-0.2,-0.2) circle (0.08cm);
\filldraw[color=black, fill=black](0.2,-0.2) circle (0.08cm);

\node (2) at (2,0) [simplecircle] {};
\filldraw[color=black, fill=black](2 + 0.2,0.2) circle (0.12cm);
\filldraw[color=black, fill=black](2+-0.2,0.2) circle (0.07cm);
\filldraw[color=black, fill=black](2+-0.05,-0.05) circle (0.1cm);
\filldraw[color=black, fill=black](2+-0.2,-0.2) circle (0.01cm);
\filldraw[color=black, fill=black](2+0.2,-0.2) circle (0.05cm);

\node (3) at (4,0) [simplecircle, draw=white] {$\ldots$};
\node (4) at (6,0) [simplecircle] {};
\filldraw[color=black, fill=black](6 + 0.1,0.1) circle (0.2cm);
\filldraw[color=black, fill=black](6+-0.23,0.1) circle (0.01cm);
\filldraw[color=black, fill=black](6+-0.12,-0.12) circle (0.01cm);
\filldraw[color=black, fill=black](6+-0.2,-0.2) circle (0.01cm);
\filldraw[color=black, fill=black](6+0.1,-0.23) circle (0.01cm);

\draw[->] (1) -- (2) node[midway,above] {$a_0, o_0$};
\draw[->] (2) -- (3) node[midway,above] {$a_1, o_1$};
\draw[->] (3) -- (4) node[midway,above] {$a_T, o_T$};

\end{tikzpicture}
    \caption{Particle Filter Belief Update}
    \label{fig:pf_summary}
\end{subfigure}
\par\bigskip
\begin{subfigure}[b]{\linewidth}
    \centering
    \begin{tikzpicture}

\tikzstyle{simplecircle} = [circle, minimum width=1cm, minimum height=1cm, draw=black]
\tikzstyle{roundedrectangle} = [rectangle, rounded corners, minimum width=1cm, minimum height=1cm, draw=black, fill=black!20]

\node (1) at (0,0) [] {$(a_0, o_0, \ldots a_T, o_T$)};

\node (2) at (3,0) [roundedrectangle] {cDGM};

\node (3) at (5,0) [simplecircle] {};
\filldraw[color=black, fill=black](5+ 0.2,0.2) circle (0.08cm);
\filldraw[color=black, fill=black](5+ -0.2,0.2) circle (0.08cm);
\filldraw[color=black, fill=black](5+ 0,0) circle (0.08cm);
\filldraw[color=black, fill=black](5+ -0.2,-0.2) circle (0.08cm);
\filldraw[color=black, fill=black](5+ 0.2,-0.2) circle (0.08cm);

\draw[->] (1) -- (2);
\draw[->] (2) -- (3);

\end{tikzpicture}
    \caption{cDGM Belief Update}
    \label{fig:cdgm_summary}
\end{subfigure}
\caption{Summary of proposed approach}
\end{figure}
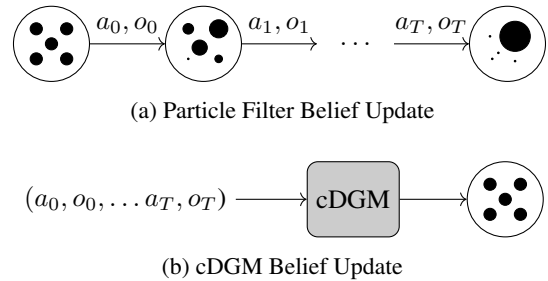

Traditional belief representations, such as particle filters, are often not suitable for high-dimensional states and large numbers of observations. They often suffer from \emph{particle depletion}, where, after successive iterations of re-weighting, only a small number of particles retain most of the weight, leading to a low quality approximation of the posterior belief (depicted in  \cref{fig:pf_summary}). Our approach (depicted in \cref{fig:cdgm_summary}) trains a deep generative model that is conditioned directly on a sequence of actions and observations, and can generate an arbitrary number of samples from the posterior belief. The cDGMs are trained from data produced by random rollout trajectories, which allows for the complex Bayesian inversion to be learned directly from the training data. 

We apply our proposed approach to solve a mineral exploration POMDP~\citep{mern2023intelligent} that has a large and continuous state space as well as continuous, noise-free observations. 
We investigate two types of cDGMs: generative adversarial networks (GANs) and denoising diffusion probabilistic models (DDPMs). We find that both model classes outperform a particle filter baseline on task-agnostic metrics, and that the DDPM models enable significanlty improved planning performance. 

Prior work~\citep{mosser2018stochastic,LALOY2017387,PhysRevE.96.043309,song2021geological} has investigated the use of deep generative models for geoscience applications, but does not condition the generation of samples on observed data. To condition to observations, traditional Bayesian inversion approaches are combined with unconditioned deep generative models. \citet{LALOY2017387} use MCMC with the generator of a variational autoencoder while \citet{liu2019deep} use a form of principle component analysis to produce conditioned samples and use deep style transfer to refine the samples. In contrast to prior work, we condition to observations directly in the generative model, and use the model for POMDP planning. 

In summary, our contributions in this paper are as follows
\begin{itemize}
\item We propose the use of conditional deep generative models to represent beliefs for belief-state planning solvers of POMDPs.
\item We experiment with a wide range of state-of-the-art deep generative models, and identity architectures that are well suited for observation conditioning. 
\item We show that cDGM belief representations outperform other representations when planning under uncertainty in a mineral exploration problem. 
\end{itemize}

\section{Background}\label{sec:background}
This section introduces the topics of POMDPs, particle filter approximations to Bayesian inference and deep generative models. Then, we provide a summary of our motivating application domain: mineral exploration.

\subsection{POMDPs}
A partially observable Markov Decision Process (POMDP) is a model for sequential decision making under state uncertainty~\citep{kochenderfer_wheeler_wray_2022}. A POMDP is defined by the tuple ($S$, $A$, $T$, $R$, $O$, $Z$, $\gamma$) where $S$ is the set of states, $A$ is the set of actions, $T$ is a transition model, $R$ is the reward function, $O$ is the observation space, $Z$ is the observation model, and $\gamma$ is the discount factor. At each timestep, an agent takes an action $a \in A$ in a state $s$ and transitions to the next state $s^\prime \sim T(\cdot \mid s, a)$, receiving a reward $r = R(s,a)$ and an observation $o \sim Z(\cdot \mid s, a, s^\prime)$. Since states are not fully observed, the agent maintains a probability distribution over states or belief $b(s)$ that is updated based on the action-observation trajectory $\tau = \{a_1, o_1, a_2, o_2, \ldots\}$. The agent's policy $a = \pi(b)$ maps beliefs to actions with the goal of maximizing the expected discounted sum of future rewards. 

POMDPs can be solved exactly only for problems with small state, action, and observation spaces, but for larger problems, approximate solvers are used~\citep{kochenderfer_wheeler_wray_2022}. The most scalable approaches often use forward tree search, where hypothetical trajectories are simulated and used to estimate the value of different actions. Some examples of tree-search-based POMDP solvers include POMCP~\citep{silver2010monte}, POMCPOW and PFT~\citep{sunberg2018online}.

\subsection{Particle Approximations for Bayesian Inference}

Accurate belief representations are an essential element to solving POMDPs, both for planning and execution. After taking action $a$, the prior belief $b(s)$ is updated with the observation $o$ using Bayes' rule
\begin{equation}
    b(s^\prime) = O(o \mid a, s^\prime) \int_{s} T(s^\prime \mid s, a) b(s) ds
\end{equation}
Exact Bayesian inference is often intractable for nonlinear dynamics, and beliefs that are not well represented by simple parametric models. Instead, the belief can be approximated by a weighted set of state samples, or particles. The weight of each particle is computed as the conditional likelihood of the state give the action-observation trajectory $w = p(s \mid \tau)$. To avoid retaining $\tau$, implementations often use a bootstrap filtering algorithm where the belief is represented by $N$ equally weighted samples of the state $b \approx \{s_i\}_{i=1}^N$ that are updated according to the following computational steps:
\begin{enumerate}
    \item Sample new states from the transition function: $s_i^\prime \sim T(\cdot \mid s_i, a)$.
    \item Compute particle weights according to the observation model: $w_i = O(o \mid a, s_i^\prime)$.
    \item Resample $N$ particles from $\{s_i^\prime\}_{i=1}^N$ according to their weights$\{w_i\}_{i=1}^N$. 
\end{enumerate}

Particle filters are flexible belief representations which have favorable convergence properties~\citep{crisan2002survey}, but they require the observation likelihood $O(\cdot \mid, a, s^\prime)$. In problems where the observation likelihood is undefined (as with noiseless observations), or intractable to compute, an additional approximation is needed to compute the likelihood weights. Approximate Bayesian computation (ABC) is a particle filtering approach where the likelihood function is replaced with a approximation $\tilde{O}$. For example, in the case of noiseless observations where $o=Z(s)$, an approximate observation model could be a Gaussian of the form 
\begin{equation}
    \tilde{O}(o \mid a, s_i) \propto e^{-\| o - Z(s) \|^2 / 2\sigma_{\rm ABC}^2}
    \label{eq:abc}
\end{equation}
for a user-specified value of $\sigma_{\rm ABC}$. ABC introduces bias into the belief representation, but is often found to be a useful approximation in practice~\citep{csillery2010approximate}.

Particle approximations to the belief can suffer from particle depletion, where only a small number of particles retain most of the weight. Particle depletion is exacerbated when the state space is high dimensional and where there are long sequences of actions and observations~\citep{snyder2008obstacles}. It may be mitigated through particle injection, where new particles matching current actions and observations are added to the belief representation~\citep{van2019particle}. The tractability of adding new particles, however, depends on the efficiency of generating new state samples. 

\subsection{Deep Generative Models}
Deep generative models (DGMs) are machine learning models that are trained to represent the underlying distribution of training data, $x \sim P_x$, enabling the generation of novel samples that closely resemble the original data. Well-trained generative models not only capture the salient features of the training data, but also facilitate the generation of diverse outputs. Due to the complexity of the training data distribution, deep neural networks (DNNs) are often used as the model class, but the training approaches can vary widely. In this study, we use two prominent types of DGMs, namely generative adversarial networks (GANs) and denoising diffusion probabilistic models (DDPMs).

\paragraph{Generative Adversarial Networks (GANs)}
GANs consist of two components: the generator network $G$ which generates new samples $\tilde{x}$, and the discriminator network $D$, which learns to distinguish the real and generated data. The two networks are trained simultaneously, competing against each other in a minimax game. A GAN has converged once the discriminator is unable to distinguish the generated and real data. Despite their many recent success~\citep{sauer2022stylegan,kang2021rebooting}, GANs often lack stability during training leading to low-quality samples, or mode collapse, where the diversity of samples is compromised
~\citep{kynkaanniemi2019improved,saxena2021generative}.

Several loss functions have been proposed for training GANs. Let $z \sim P_z$ be a latent vector sampled from a known probability distribution. The binary cross entropy loss~\citep{goodfellow2014generative} treats the discriminator as a classifier model where real images that are labeled $1$ and generated images are labeled $0$, while the generator is trained to maximize the discriminator's loss:
\begin{align}
    L_{D} &= \mathbb{E}_{\substack{x \sim P_x\\z \sim P_z}} \left[ \log(D(x)) +  \log(1-D(G(z))) \right]\\
    L_{G} &= \mathbb{E}_{z \sim P_z}\left[\log(1-D(G(z))) \right]
\end{align}
The Wasserstein loss~\citep{arjovsky2017wasserstein} was proposed as a way to stabilize training and is given by
\begin{align}
    L_{D} &= \mathbb{E}_{\substack{x \sim P_x\\z \sim P_z}}\left[ -D(x) + D(G(z)) \right]\\
    L_{G} &= \mathbb{E}_{z \sim P_z}\left[ -D(G(z)) \right]
\end{align}
The Wasserstein loss is often coupled with a gradient penalty, which limits the norm of the gradient when training the discriminator~\citep{gulrajani2017improved}.

For our experiments, we consider two types of DNN architectures.
\begin{enumerate}
    \item Deep Convolutional GAN (DCGAN): The DCGAN architecture uses feed-forward convolutional neural networks for both the generator and the discriminator. The specific number of convolutional filters and layers may vary, but our architecture is based on prior work~\citep{mirza2014conditional}.
    \item StyleGAN: The StyleGAN architecture was proposed by~\citet{karras2019style} and obtained state-of-the art-performance on image generation tasks. Contrary to traditional approaches that feed the latent vector only into the first layer, the StyleGAN generator maps the latent vector to an intermediate space, which is used in each layer of the network.
\end{enumerate}

\paragraph{Denoising Diffusion Probabilistic Models (DDPMs)}
DDPMs are based on the idea of using a diffusion process to iteratively de-noise a noisy input and generate a clean output. During training, a random $t \in [1, T]$ is sampled, where $T$ is the number of denoising steps. Given the original image $x_0$, $t$-steps of Gaussian noise can be added by computing
\begin{equation}
    x_t = \sqrt{\overline{\alpha}_t}x_0 + \sqrt{1-\overline{\alpha}_t}\epsilon
\end{equation}
where $\epsilon \sim \mathcal{N}(0,I)$ is Gaussian noise of the same dimension as $x$ and $\overline{\alpha}_t = \prod_{s=1}^t \alpha_s$ for a noise schedule $\alpha_t$. A network $U$ (typically a UNet model~\citep{ronneberger2015u}) is trained using the mean squared error between its prediction of the image noise and the actual noise
\begin{equation}
    L = \|U(x_t, t)-\epsilon\|^2
\end{equation}

During inference, we start at step $T$, with $x_T \sim \mathcal{N}(0,I)$, and iteratively compute $x_{t-1}$ with the following recurrence relation:
\begin{equation}
    x_{t-1} = \frac{1}{\alpha_t}\left(x_t - \frac{1-\alpha_t}{\sqrt{1-\overline{\alpha}_t}}U(x_t, t)\right) + (1-\alpha_t)\epsilon \label{eq:ddpm_denoise}
\end{equation}
where $\epsilon \sim \mathcal{N}(0,1)$ when $t > 1$ and $\epsilon = 0$ otherwise. See \citet{ho2020denoising} for more details about the UNet architecture and the (de)noising process.

DDPMs are known for both the quality of their outputs and their training stability. However, the major drawback is the time of inference. In order to produce a sample, DDPMs require $T$ inference steps where $T$ is usually between \num{100} and \num{1000}. In contrast, GANs only require a single inference pass to produce a sample.

\subsection{Mineral Exploration POMDP}
Batteries are a key technology enabling the transition from fossil fuels to renewable forms of energy, but require large amount of rare earth elements such as copper, nickel, cobalt and lithium~\citep{sovacool2020sustainable}. The rate of discovery of new mineral deposits has decreased while the demand for these elements has increased~\citep{davies2021learning}. Mining operations for these elements are expensive and can have significant environmental and social impacts, so finding ore deposits that are economically viable to mine and in the right geographical regions is essential. 

Recent work has formulated the mineral exploration problem as a POMDP~\citep{mern2023intelligent}. The states are \num{2}-dimensional $32 \times 32$ maps (see \cref{fig:samples}), with each grid cell having a real value in $[0,1]$ indicating the local ore concentration. The actions are to drill observational boreholes, or make a final decision to abandon the project or go forward with the mining operation and extract the ore. When a borehole is drilled, the ore density at the drill location is observed directly without any noise. The cost of drilling boreholes is a fixed value $R_{\rm drill}(s) = -0.1$, and we fix a discrete set of potential borehole locations on a $\num{6} \times \num{6}$ grid. If the project is abandoned, then the episode ends with no additional reward $R_{\rm abandon}(s) = 0$, but if the agent chooses to mine, then the reward is the sum of grid cells where the ore quantity is above a concentration threshold $\rho = 0.7$ minus a fixed capital expenditure $R_{\rm capex} = -52$, 
\begin{equation}
R_{\rm mine}(s) = R_{\rm capex} + \sum_{i=1}^{32}\sum_{j=1}^{32} \mathbbm{1}_{\{ s_{i,j} > \rho\}} 
\end{equation}
where $\mathbbm{1}$ is the indicator function and $s_{i,j}$ is the ore value at row $i$ and column $j$. There are no dynamics to this problem, making it a POMDP-lite~\citep{chen2016pomdp}, where information-gathering is the primary goal.

Each mineral deposit is defined by a \num{100}-point polygon that is smoothed with a Gaussian kernel and has Gaussian process noise added on top. The initial state distribution is defined and sampled from using a publicly available repository\footnote{\smaller \url{https://github.com/sisl/MineralExploration}}. The values of $\rho$ and $R_{\rm capex}$ were selected so that the mean return of the initial state distribution is near \num{0}.

\section{Conditional Deep Generative Models} \label{sec:cDGM}
This section discusses the conditioning of deep generative models and outlines the design choices made for both GANs and DDPMs. We then introduce a variety a metrics that can be used to evaluate conditional generative models and discuss how they can be incorporated into a POMDP solver.

\subsection{Conditioning the models}

The goal of conditional generative models is to represent the conditional probability distribution $P_{x \mid c}$ instead of the marginal distribution $P_x$, where $c$ is a condition that describes $x$~\citep{mirza2014conditional,ho2020denoising}. In image generation, the conditioning is often done based on a desired class from a discrete set (e.g. conditioning on the digit in MNIST generation). To train a class-conditional generative model, the class label (typically as a one-hot encoding) is passed to the DNN (often after being further encoded via a trainable embedding layer) along with the corresponding training data point. Conditioning on continuous values has been studied in the domain of super-resolution generative models~\citep{ledig2017photo} as well as image in-painting~\citep{yu2018generative}. The continuous condition is encoded in a matrix that contains data about all or part of the pixels of the image and is concatenated with the latent vector (with or without some additional encoding)~\citep{ledig2017photo,yu2018generative}.

The mineral exploration problem has continuous conditioning variables (the mineral density observations), but unlike prior work on super-resolution and in-painting, the conditioning variables are sparse: between $1$ and $36$ points on a $32\times32$ grid. The mineral density observation is encoded into a tensor $c$ of shape $2\times32\times32$ where the first channel is a one-hot encoding of where the observations have been made and the second channel contains the corresponding observation value, and is \num{0} anywhere that has not been observed. 

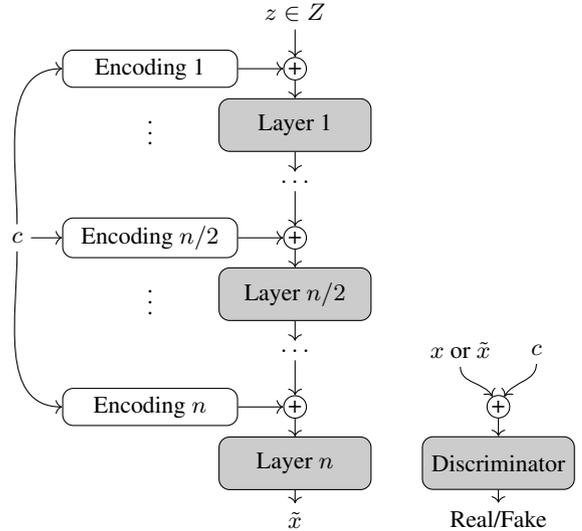
\begin{figure}[t]
    \centering
    \begin{subfigure}[b]{0.7\linewidth}
        \centering
        \begin{tikzpicture}
\small
    \tikzstyle{layer_rectangle}=[draw=black, shape=rectangle, rounded corners, minimum width=2cm, minimum height=0.7cm, fill=black!20]
    \tikzstyle{encoding_rectangle}=[draw=black, shape=rectangle, rounded corners, minimum width=2.3cm]
    \tikzstyle{concat_node}=[fill=white, draw=black, shape=circle]
    \tikzset{node distance = 0.75cm and 1cm}
    
    \node                                   (latent) at (0, 0) {$z \in Z$};
    \node [concat_node, below of=latent,]    (c1) {};
    \node at (c1) {+};
    \node [layer_rectangle, below of=c1]    (layer1) {Layer 1};
    \node [, below of=layer1]               (v1) {$\hdots$};
    \node [concat_node, below of=v1]        (cnb2) {};
    \node at (cnb2) {+};
    \node [layer_rectangle, below of=cnb2] (layernb2) {Layer $n/2$};
    \node [below of=layernb2]              (v2) {$\hdots$};
    \node [concat_node, below of=v2]        (cn) {};
    \node at (cn) {+};
    \node [layer_rectangle, below of=cn]   (layern) {Layer $n$};
    \node [below of=layern]   (x) {$\tilde{x}$};

    \draw [->] (latent) -- (c1);
    \draw [->] (c1) -- (layer1);
    \draw [->] (layer1) -- (v1);
    \draw [->] (v1) -- (cnb2);
    \draw [->] (cnb2) -- (layernb2);
    \draw [->] (layernb2) -- (v2);
    \draw [->] (v2) -- (cn);
    \draw [->] (cn) -- (layern);
    \draw [->] (layern) -- (x);

    \node [encoding_rectangle,  left of=c1, anchor=east]  (e1) {Encoding 1};
    \node [below of=e1] {$\vdots$};
    \node [encoding_rectangle, left of=cnb2, anchor=east]  (enb2) {Encoding $n/2$};
    \node [below of=enb2] {$\vdots$};
    \node [encoding_rectangle, left of=cn, anchor=east]  (en) {Encoding $n$};

    \draw [->] (e1) -- (c1);
    \draw [->] (enb2) -- (cnb2);
    \draw [->] (en) -- (cn);

    \node [left of=enb2, xshift=-1cm] (condition) {$c$};
    \draw [->, in=-180, out=90] (condition) to (e1);
    \draw [->] (condition) -- (enb2);
    \draw [->, in=180, out=-90] (condition) to (en);

\end{tikzpicture}
        \caption{Generator} \label{fig:generator_conditioning}
    \end{subfigure}\hfill
    \begin{subfigure}[b]{0.3\linewidth}
        \centering
        \begin{tikzpicture}
\small
    \tikzstyle{layer_rectangle}=[draw=black, shape=rectangle, rounded corners, minimum width=2cm, minimum height=0.7cm, fill=black!20]
    \tikzstyle{concat_node}=[fill=white, draw=black, shape=circle]
    \tikzset{node distance = 0.75cm and 0.5cm}
    
    \node [concat_node] (c1) at (0,0) {};
    \node at (c1) {+};
    \node [layer_rectangle, below of=c1]    (d) {Discriminator};
    \node [below of=d]               (out) {Real/Fake};
    \node [above of= c1, xshift=-0.5cm] (in) {$x$ or $\tilde{x}$};
    \node [above of= c1, xshift=0.5cm] (c) {$c$};

    \draw [->, out=-90, in=60] (c) to (c1);
    \draw [->, out=-90, in=120] (in) to (c1);
    \draw [->] (c1) -- (d);
    \draw [->] (d) -- (out);
    
\end{tikzpicture}
        \caption{Discriminator} \label{fig:discriminator_conditioning}
    \end{subfigure}
    \caption{GAN Conditioning}
    \label{fig:gan_conditioning}
\end{figure}

\begin{figure}[t]
    \centering
    \begin{tikzpicture}
\small
    \tikzstyle{layer_rectangle}=[draw=black, shape=rectangle, rounded corners, minimum width=2cm, minimum height=0.7cm, fill=black!20]
    \tikzstyle{concat_node}=[fill=white, draw=black, shape=circle]
    \tikzset{node distance = 0.75cm and 0.75cm}
    
    \node [concat_node] (c1) {};
    \node at (c1) {+};
    \node [layer_rectangle, below of=c1]    (d) {UNet};
    \node [layer_rectangle, below of=d, yshift=-0.3cm]  (out1) {\cref{eq:ddpm_denoise}};
    \node [below of=out1, yshift=-0.2cm]               (out) {De-noised Sample};
    \node [above left of= c1, yshift=0.2cm] (in) {Noisy Image};
    \node [above right of= c1, yshift=0.2cm] (c) {$c$};

    \draw [->, out=-90, in=60] (c) to (c1);
    \draw [->, out=-90, in=120] (in) to (c1);
    \draw [->] (c1) -- (d);
    \draw [->] (d) -- (out1);
    \draw [->] (out1) -- (out);

    \draw [->, out=-180, in=180] (out.west) to (in.west) node[midway,left, xshift=-2.4cm, yshift=-1cm] {Input at $t-1$};
    
\end{tikzpicture}
    \caption{DDPM Conditioning on step $t$ of denoising.}
    \label{fig:ddpm_conditioning}
\end{figure}
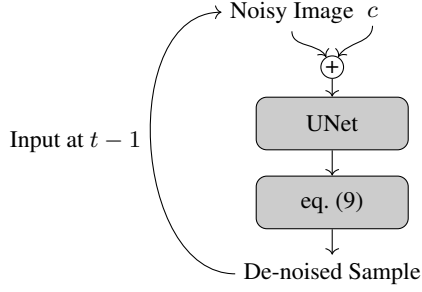

The conditioning strategies for the GAN models are shown in \cref{fig:gan_conditioning}. At the first layer the condition $c$ is passed through an encoding layer and concatenated with the latent vector of the generator, and concatenated to the given sample for the discriminator. Motivated by prior work~\citep{karras2019style} and initial experimentation, we found that injecting the condition at additional layers proved useful in producing an accurate belief representation. We therefore consider two additional configurations: \emph{half}, where we inject the condition into each of the first half of the layers of the generator and \emph{all} where it injected at every layer. Each layer that the condition is injected in has its own trainable encoding layer. A similar approach is employed for conditioning the DDPM model as shown in \cref{fig:ddpm_conditioning}. The 2-channeled condition is concatenated to each noisy image prior to its input into the network.

\subsection{Belief Evaluation Metrics} \label{sec:metrics}
Evaluating the quality of a conditional generative model is challenging for complex and high dimensional distributions. Prior work has primarily focused on image generation of common objects or human faces, and the corresponding metrics have therefore been designed to correlate with subjective notions of image realism~\citep{heusel2017gans}. In this work however, we are primarily concerned with how well the belief representation covers the training distribution and matches the observations/conditions, as well as how the belief representation leads to good planning performance. We therefore study a variety of task-agnostic and task-specific metrics.

To do so, we construct an evaluation set of $N_{\rm test}$ state-history pairs $(s_i, \tau_i)$. For each sample in the evaluation set, the posterior is modeled using the chosen belief representation and used to produce \num{500} posterior samples. From those samples we can compute the following metrics. 

\paragraph{Task-agnostic metrics}
We wish to know how well the generative model distribution covers the posterior distribution over states conditioned on a set of actions and observations. We therefore measure test set recall~\citep{kynkaanniemi2019improved} as well as a measure of observation similarity to evaluate the cDGMs in a model agnostic way. 
\begin{itemize}
    \item \emph{Test set recall}: To evaluate recall, we determine how well each belief representation can reproduce samples from the test set. We measure the distance between the closest generated sample in the belief to each of the test set points:
    \begin{equation}
    \frac{1}{N_{\rm test}}\sum_{i=1}^{N_{\rm test}} \min_{\tilde{s} \sim b(\cdot \mid \tau_i)} \| \tilde{s} - s_i \|
    \end{equation}
    Initial experimentation found that an $L_2$ norm between samples was sufficient to distinguish high quality models from low quality ones, so we use it for our distance metric. 
    \item \emph{Observation error}: To evaluate the conditioned points we measure the distance between the imposed conditioning and the resulting observations from produced samples:
    \begin{equation}
        \frac{1}{N_{\rm test}}\sum_{i=1}^{N_{\rm test}} \mathbb{E}_{\tilde{s} \sim b(\cdot \mid \tau_i)} \| Z(\tilde{s}) - Z(s_i) \|
    \end{equation}
    \item \emph{Wall-clock time}: To evaluate the computational efficiency of a belief representation, we measure the amount of time it takes to produce samples from the posterior. 
\end{itemize}

\paragraph{Task-specific metrics}
Knowing the downstream task that the belief representation will be used for allows us to design task-specific metrics for belief evaluation. For the mineral exploration problem, we can compute the expected value of the ore in the mineral deposit
\begin{equation}
    \bar{R}_i = \mathbb{E}_{\tilde{s} \sim b(\cdot \mid \tau_i)} \left[ R_{\rm mine}(\tilde{s}) \right]
\end{equation}
and to quantify uncertainty, we can compute the variance of the belief over the value of ore as
\begin{equation}
    \sigma_i^2 = \text{Var}_{\tilde{s} \sim b(\cdot \mid \tau_i)}\left[ R_{\rm mine}(\tilde{s}) \right]
\end{equation}
We then compute the following metrics:

\begin{itemize}
    \item \emph{Ore value error}: This metrics measure how well the belief models the true value of the ore. The mean value of the ore is computed from the posterior samples and we measure the distance between the estimated value and the ground truth for each sample:
    \begin{equation}
    \frac{1}{N_{\rm test}}\sum_{i=1}^{N_{\rm test}} \left| R_{\rm mine}(s_i) - \bar{R}_i \right|
    \end{equation}

    \item \emph{Decision accuracy} measures how well the belief informs the final mine/abandon decision of the agent. The agent is forced to make a mine or abandon decision based only on the posterior samples. The decision to mine is correct if there is a positive economic return for mining, and the decision to abandon correct if there is a negative economic return. The decision accuracy is computed as
    \begin{equation}
        \frac{1}{N_{\rm test}}\sum_{i=1}^{N_{\rm test}} \mathbbm{1}_{\left\{\text{sign}\left(R_{\rm mine}(s_i)\right) = \text{sign}\left(\bar{R}_i\right)\right\}}
    \end{equation}
    
    \item \emph{Probability density of ore value} measures how well the belief representation models uncertainty. We measure the probability density of the ground truth for each sample, assuming a normal probability distribution for the estimated ore value. 
    \begin{equation}
        \frac{1}{N_{\rm test}}\sum_{i=1}^{N_{\rm test}} \mathcal{N}(R_{\rm mine}(\tilde{s}) \mid \bar{R}_{i}, \sigma_i^2)
    \end{equation}
    where $\mathcal{N}$ is a normal distribution.
\end{itemize}

\subsection{Planning with {c}DGMs}

Beliefs represented by cDGMs can replace other implicit belief representations used in POMDP planning algorithms. For example, the particle filter tree (PFT) algorithm~\citep{sunberg2018online} uses Monte Carlo tree search (MCTS) over the belief space approximated with particles, and uses a particle filter to add new belief nodes to the tree. A cDGM could replace this representation directly, however, MCTS is an inherently serial algorithm, requiring only one new posterior sample at each iteration. To get the full benefit of GPU hardware acceleration, we want an algorithm that can batch the generation of posterior samples, so in this work we devise a POMDP solver based on value of information (VOI)~\citep{kochenderfer2019algorithms}, which easily parallelizes the generation of posterior samples. While a policy based on VOI will not recover the optimal POMDP solution, in practice it produces competitive results (see \cref{tab:planning_results}). Also, the VOI policy explicitly reasons about information gain, which better differentiates the quality of the different belief representations. 

\begin{algorithm}
\caption{Value of Information Policy} \label{alg:voi}
\begin{algorithmic}[1]
    \Function{VOIPolicy}{$\mathcal{P}$, $b$}
    \State $\{s_i\}_{i=1}^N \sim b$ \label{line:sample}
    \State $V \gets \emptyset$ \label{line:Vinit}
    \For{$a$ in $A$}
        \State $V_a \gets \emptyset$
        \For{$i$ in $\{1, \ldots, N\}$}
            \State $s^\prime_i, o_i, r_i \gets \text{gen}(\mathcal{P}, s_i, a)$ \label{line:gen}
            \State $b^\prime_i \gets \text{update}(b, a, o_i)$ \label{line:update}
            \State $\{ s_j \}_{j=1}^M \sim b^\prime_i$ \label{line:posterior_samples}
            \State $\bar{R}_i \gets \text{mean}(\{ R_{\rm mine}(s_j) \}_{j=1}^M)$ \label{line:meanval}
            \State $V_a \gets V_a \cup r_i + \max(0, \bar{R}_i)$ \label{line:val_w_decision}
        \EndFor
        \State $V \gets V \cup \text{mean}(V_a)$ \label{line:a_avg}
    \EndFor
    \State \Return{$A[\text{arg max} \ V_{\rm drills}]$} \label{line:return}
    \EndFunction
\end{algorithmic}
\end{algorithm}

The algorithm (see \cref{alg:voi}) takes as input the POMDP $\mathcal{P}$ and the current belief $b$ and returns the next action. The algorithm starts by taking $N$ samples from the belief (line \ref{line:sample}) and initializes an array of values (line \ref{line:Vinit}) that will correspond to each action in the action space $A$. Then, for each action, we loop over the sampled states and for each one, sample a next state, observation, and reward from the POMDP (line \ref{line:gen}). The observation is used to update the belief (line \ref{line:update}) from which $M$ new states are sampled (line \ref{line:posterior_samples}). Those samples are used to compute the estimated value of the belief (line \ref{line:meanval}). The value for action $a$ in state $s_i$ is computed by assuming the next action is a terminal one (line \ref{line:val_w_decision}). The expected value for action $a$ is computed by averaging over the $N$ sampled states (line \ref{line:a_avg}), and the action with the highest value is returned (line \ref{line:return}).

\section{Experiments}


We trained a variety of cDGM models on training data consisting of \num{90000} ore map samples. The action-observation histories were sampled randomly for each batch of training data. The number of actions $N_a$ was sample from an exponential distribution $N_a \sim 1+ \text{Exp}(\lambda = 0.2)$ and the drill locations were sampled uniformly at random from the $32 \times 32$ grid. We trained our models for \num{50} epochs, using a learning rate of \num{4e-4} for the generator and the discriminator in the GANs and \num{2e-4} for the DDPMs.

For the GAN models we experimented with the injection strategy (1st, half, all),  the number of channels in the convolutions $N_{\rm channels} \in \{2,8\}$, the dimension of the latent vector $N_z \in \{32, 64, 128 \}$, and the loss function (Wasserstein (W), cross-entropy (CE), and Wasserstein with gradient penalty (W-GP). The GAN models are denoted ``GAN (injection, $N_{\rm channels}$, $N_z$, loss)". For the DDPM models we experimented with model size (small, medium, large) as well as the number of denoising iterations $N_{\rm iterations} \in [100, 1000]$ and denote these models ``DDPM (size, $N_{\rm iterations}$)". We compared cDGM belief representations to ABC particle filters with the approximate likelihood given in \cref{eq:abc}. For the particle filter beliefs, we vary the number of particles ($N_{\rm Particles} \in \{1\text{k}, 10\text{k}, 100\text{k} \}$) and choice of likelihood width ($\sigma_{\rm ABC} \in \{ 0.01, 0.05, 0.1\}$). The particle filter baselines are denoted ``PF ($N_{\rm Particles}$, $\sigma_{\rm ABC}$)". Additional details and experiments are reported in the supplemental material. In the following analysis we selected the highest performing belief representations from each model type. 

\subsection{Evaluating Belief Representations}
\begin{figure*}
    \centering
    \includegraphics[width=0.7\textwidth]{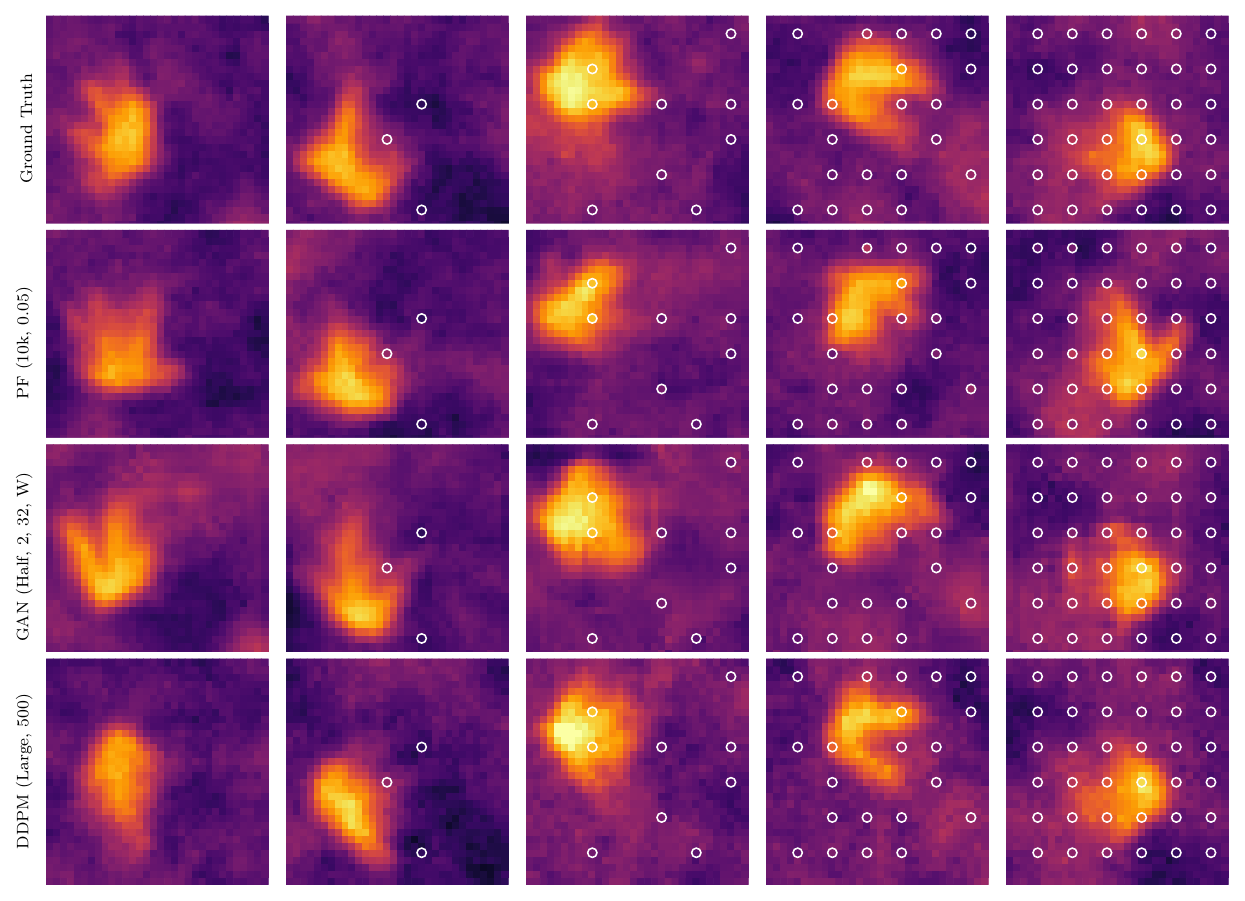}
    \caption{Samples generated from each belief representation. The top row consists of different ground-truth samples, and each other row contains the closest generated sample (out of \num{500}) for each belief representation. The white circles which drilling locations and were used for conditioning the sample. The color inside the white circle shows the ground truth observation.}    
    \label{fig:samples}
\end{figure*}

We first evaluate each belief representation via a qualitative analysis of generated samples (shown in \cref{fig:samples}). We see that while all samples are able to roughly match the size, shape and location of the ore deposit, there are minor variations that could be important for POMDP planning. Due to particle depletion, the best particle filter samples don't match the fine details of the ground truth sample, and can't improve with additional observations once the particle set is down to a single sample. The GAN model matches the shape of the ore deposit but has artifacts in the space between observations, leading to a lower predicted value of ore in the deposit. The DDPM model is able to match the ore deposit closely, and has the closest agreement with the ground truth with the maximum number of observations. 

\begin{table*}[ht]
    \caption{Comparison of belief representations across task-agnostic and task-dependent evaluation metrics. Reporting mean and standard deviation in parentheses.}
    \label{tab:belief_evaluation}
    \centering
    \begin{tabular}{@{}lcccccc@{}} 
    \toprule
\makecell{Belief\\Representation} & \makecell{Min\\$L_{2}$($\downarrow$)} & \makecell{Conditioning\\Error($\downarrow$)} & \makecell{Ore Value\\Error($\downarrow$)} & \makecell{Decision\\Accuracy($\uparrow$)} & \makecell{Prob. of\\Ore Value($\uparrow$)} & \makecell{Time($\downarrow$)}\\
\midrule
PF ($10$k, $0.05$) & $2.74 (0.30)$ & $0.05 (0.01)$ & $14.39 (11.33)$ & $0.67 (0.47)$ & $0.02 (0.03)$ & $0.36 (0.21)$\\
PF ($100$k, $0.05$) & $2.52 (0.28)$ & $0.04 (0.01)$ & $13.35 (10.44)$ & $0.70 (0.46)$ & $0.02 (0.03)$ & $3.93 (2.20)$\\
GAN (Half, $2$, $32$, W) & $2.47 (0.56)$ & $0.03 (0.01)$ & $15.90 (13.71)$ & $0.70 (0.46)$ & $0.02 (0.02)$ & $\mathBF{0.22 (0.00)}$\\
GAN (All, $2$, $128$, W) & $2.43 (0.51)$ & $0.03 (0.01)$ & $20.01 (16.61)$ & $0.62 (0.49)$ & $0.02 (0.02)$ & $0.23 (0.00)$\\
DDPM (Large, $250$) & $2.04 (0.49)$ & $\mathBF{0.01 (0.00)}$ & $12.72 (11.24)$ & $0.71 (0.45)$ & $0.02 (0.01)$ & $29.17 (0.02)$\\
DDPM (Large, $500$) & $\mathBF{2.02 (0.47)}$ & $\mathBF{0.01 (0.00)}$ & $\mathBF{12.18 (10.25)}$ & $\mathBF{0.72 (0.45)}$ & $0.02 (0.01)$ & $58.56 (0.03)$\\
    \bottomrule
    \end{tabular}
\end{table*}

We make these qualitative observations more concrete by evaluating each belief representation according to the metrics described in \cref{sec:metrics} with a test set with $N_{\rm test} = 1300$. The results are reported in \cref{tab:belief_evaluation}. Across almost all of the metrics the DDPM models outperformed the other belief representations, but at the cost of higher wall-clock time. The GAN models performed well on the task-agnostic metrics, but suffered in the task-specific metrics, though they were by far the most computationally efficient representation. 
    
\begin{figure*}
    \centering
    \input{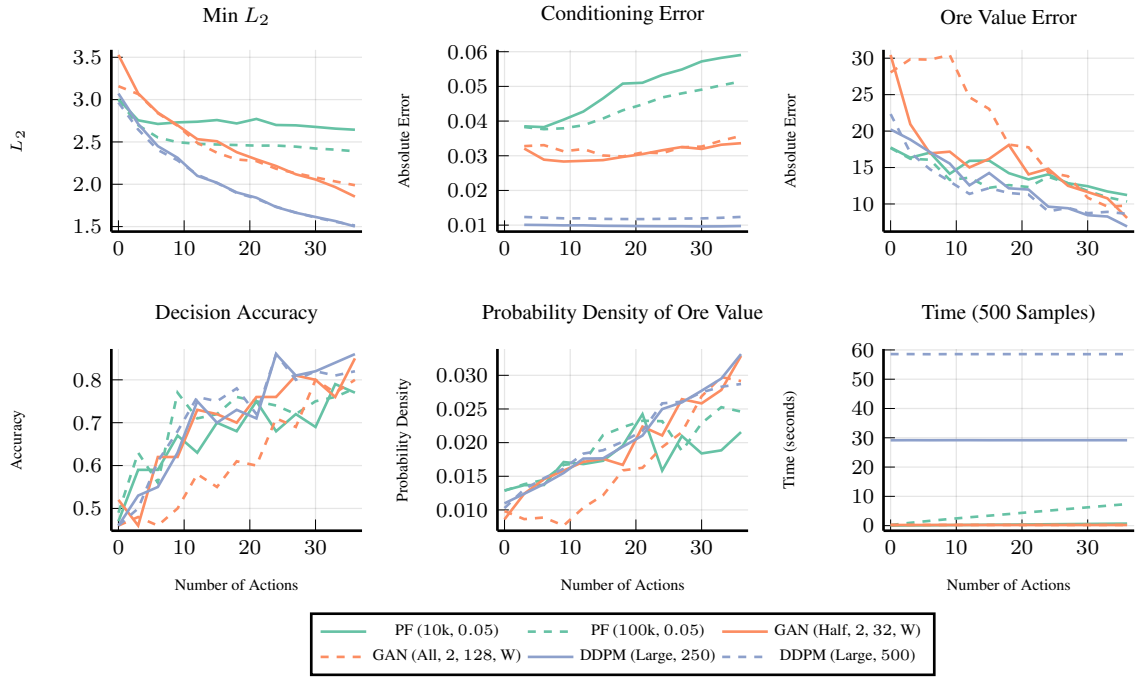}
    \caption{Belief metrics vs. number of actions/observations.}
    \label{fig:beliefComparisons}
\end{figure*}

In addition to studying the average value of the metrics, we also investigate each metric versus the number of actions and observations used to condition the posterior and plot the results in \cref{fig:beliefComparisons}. The cDGMs continue to reduce the minimum $L_2$ distance with more observations (with DDPMs outperforming GANs) while the particle filters flatten out after about \num{10}, due to particle depletion. Both the particle filter and to a lesser extent, the GANs have increased average observation error with the number of observations while the DDPM models have a consistently low observation error. Over value error decreases and decision accuracy increases with number of observations, but the GAN models are typically lower performing. The probability density of the ground truth increases with observations (as the posterior belief becomes narrower), but this effect is limited for the particle filters, again due to particle depletion. Lastly, we note that the wall clock time for the particle filter representation scales linearly with the number of observations, while the cDGMs have a constant computational cost, albeit much larger for the DDPM models.

\subsection{Planning Experiments}

\begin{table}
    \caption{Planning performance with various belief representations.} \label{tab:planning_results}
    \centering
    \begin{tabular}{@{}lc@{}} 
    \toprule
    POMDP Solver & Return \\
    \midrule
    POMCPOW w/ PF ($10$k, $0.05$) & $4.44$ \\
    VOI w/ PF ($10$k, $0.05$) & $3.37$ \\
    VOI w/ PF ($10$k, $0.1$) & $6.10$ \\
    VOI w/ GAN (Half, $2$, $32$, W) & $1.78$\\
    VOI w/ GAN (All, $2$, $128$, W) & $2.45$\\
    VOI w/ DDPM (Large, $250$) & $6.42$ \\
    VOI w/ DDPM (Large, $500$) & $\mathbf{7.04}$\\
    \bottomrule
    \end{tabular}
\end{table}

For the VOI policy we use $N=50$ observations and $M=10$ samples to estimate the return. Additionally, we consider a set of $|A| = 50$ actions that include multiple bore-hole locations. The actions are sampled by first selecting the number of bore-hole locations (uniform in $[1,10]$) and then sampling that number of undrilled locations on the $6 \time 6$ grid. The results of our planning experiments (evaluated on $46$ test ore maps) are shown in \cref{tab:planning_results}. The VOI policy with DDPM models obtains the highest return with the \num{500}-iteration model performing better than the \num{250}-iteration model. They outperform the POMCPOW algorithm with a particle filter belief as well as the baseline particle filters used with the same VOI policy. The GAN models perform comparatively poorly, as predicted by the task-dependent measures of the belief representation.

\section{Conclusion}
In this work we found that cDGMs can indeed be useful representations of the belief for belief state planning, even when compared to a particle filter with a similar number of samples used to train the cDGM. We found, however, that the specific algorithm used to train the belief is important to planning performance (i.e. DDPMs significantly outperform GANs). The difference between the quality of belief representations should be evaluated with task-specific metrics, as task agnostic metrics may be misleading (as shown by GANs outperforming particle filters on the task agnostic metrics but ultimately having worse planning performance). 

\paragraph{Limitations and Future Work}
Our work is subject to a number of limitations that we plan to address in future work:
\begin{itemize}
    \item The current POMDP does not have a dynamics function, which makes it easier to process sequences of actions and observations. Future work will study POMDPs with a dynamics function, and apply time-series modeling approaches to embedding the action-observation sequence into a fixed sized condition vector.
    \item The mineral exploration POMDP currently relies on synthetic data that lies on a relatively low-dimensional manifold. Future work will use higher-fidelity geological data, a setting where particle filtering performs even worse. 
    \item We only compare our cDGM belief representations to particle filtering approaches. More advanced belief representations have been developed for geological data and could be compared to our cDGM approach. 
\end{itemize}



\begin{contributions} 
    Briefly list author contributions. 
    This is a nice way of making clear who did what and to give proper credit.
    This section is optional.

    H.~Q.~Bovik conceived the idea and wrote the paper.
    Coauthor One created the code.
    Coauthor Two created the figures.
\end{contributions}

\begin{acknowledgements} 
    Briefly acknowledge people and organizations here.

    \emph{All} acknowledgements go in this section.
\end{acknowledgements}

\bibliography{references.bib}
\end{document}